\documentclass{article}

\usepackage[final]{neurips_2024}

\usepackage{hyperref}
\usepackage{url}
\usepackage{graphicx}
\usepackage{subcaption}
\usepackage[dvipsnames,table]{xcolor}
\usepackage{multirow}
\usepackage{booktabs}
\usepackage{natbib}
\usepackage{amsmath} 
\usepackage{amssymb}
\usepackage{tabularx}
\usepackage{enumitem}
\usepackage[capitalize,noabbrev]{cleveref}
\hypersetup{
    colorlinks=true,
    bookmarksnumbered,
    pdfstartview={FitH},
    linktoc=all,
    citecolor=Turquoise,
    linkcolor=PineGreen,
    urlcolor=NavyBlue,
    pdfpagemode={UseOutlines}}

\newcommand{\bR}{\mathbb{R}}
\newcommand{\hH}{\bar{\mathbf{H}}}
\newcommand{\hh}{\bar{\mathbf{h}}}
\newcommand{\given}{\mid}
\newcommand{\bA}{\mathbf{A}}
\newcommand{\bY}{\mathbf{Y}}
\newcommand{\bX}{\mathbf{X}}
\newcommand{\by}{\mathbf{y}}

\title{Uncertainty-Aware Optimal Treatment Selection \\for Clinical Time Series}

\author{%
  Thomas Schwarz \\
  Technical University of Munich\\
  Helmholtz Munich
  \And
  Cecilia Casolo \\
  Technical University of Munich \\
  Helmholtz Munich \\
  Munich Center for Machine Learning (MCML)
  \And
  Niki Kilbertus \\
  Technical University of Munich \\
  Helmholtz Munich \\
  Munich Center for Machine Learning (MCML)
}

\begin{document}

\maketitle

\begin{abstract}
In personalized medicine, the ability to predict and optimize treatment outcomes across various time frames is essential. Additionally, the ability to select cost-effective treatments within specific budget constraints is critical. Despite recent advancements in estimating counterfactual trajectories, a direct link to optimal treatment selection based on these estimates is missing. This paper introduces a novel method integrating counterfactual estimation techniques and uncertainty quantification to recommend personalized treatment plans adhering to predefined cost constraints. Our approach is distinctive in its handling of continuous treatment variables and its incorporation of uncertainty quantification to improve prediction reliability. We validate our method using two simulated datasets, one focused on the cardiovascular system and the other on COVID-19. Our findings indicate that our method has robust performance across different counterfactual estimation baselines, showing that introducing uncertainty quantification in these settings helps the current baselines in finding more reliable and accurate treatment selection. The robustness of our method across various settings highlights its potential for broad applicability in personalized healthcare solutions.
\end{abstract}

\section{Introduction} 

In recent years, there has been a growing interest within medical research in forecasting patient data across various time periods and predicting future treatment effects, a trend that aligns well with the longitudinal nature of healthcare data \citep{allam2021analyzing,liu2018estimating,feuerriegel2024causal}. This is particularly important in personalized medicine, where the goal is usually to develop treatment plans that are tailored to the predicted health development of individual patients \citep{moodie2007demystifying, van2020prediction}. A critical aspect of these plans is the reliability of the predictions \citep{utomo2018treatment, hess2023bayesian}, as practitioners often prefer more certain outcomes over those that are merely optimistic or cost-effective \citep{kerr2008role, west2002clinical}. Additionally, the cost and potential side effects of treatments must be considered, particularly for high-dose treatments \citep{weiting2022clinical, auger2005risks, frei1980dose}. 

While several studies have explored counterfactual estimations and uncertainty quantification in longitudinal data \citep{Bica2020Estimatingadversariallybalancedrepr, hess2023bayesian, gnet2021, debrouwer2022predictingimpacttreatmentstime}, there remains a gap in applying these techniques to inform treatment assignment strategies. Notably, existing strategies often overlook the continuous nature of many treatment decisions, such as chemotherapy dosing \citep{kallus2018policy, kreif2015evaluation}.

\paragraph{Contributions.} This work introduces a model-agnostic framework for optimal treatment selection that integrates uncertainty quantification of counterfactual predictions within predefined cost constraints. Our approach effectively accommodates continuous-valued treatments and is compatible with various uncertainty quantification and counterfactual prediction methodologies, enhancing its applicability across different settings.

\section{Related Work}

\paragraph{Treatment effect trajectories estimation.} In the realm of longitudinal data, recent years have seen various developments in estimating treatment effects, with a focus on the trajectories of individual patients. \citet{Bica2020Estimatingadversariallybalancedrepr} introduced a counterfactual recurrent neural network (\texttt{CRN}) that utilizes an encoder-decoder style long-short-term memory network. Following this, \citet{melnychuk22a} incorporated the transformer architecture to create the causal transformer (\texttt{CT}), and \citet{seedat22b} addressed irregularly sampled time series by employing neural controlled differential equations (\texttt{TE-CDE}). Counterfactual ODE (\texttt{CF-ODE}) \citep{debrouwer2022predictingimpacttreatmentstime} and Bayesian neural controlled differential equation \texttt{BNCDE} \citep{hess2023bayesian} later followed, including uncertainty quantification into counterfactual trajectories estimation. Recent models for counterfactual predictions based on g-computations were also introduced, like \texttt{G-NET} \citep{gnet2021}. However, these studies primarily concentrate on scenarios involving treatments with discrete values, such as fixed dosages per treatment type.

\paragraph{Balancing representations.}
Prior works have concentrated on estimating counterfactual trajectories with temporal confounding, where past covariates influence both the outcome and the treatment. To counteract treatment selection bias in time series models, \citet{Bica2020Estimatingadversariallybalancedrepr}, \citet{seedat22b} and \citet{melnychuk22a} have explored the development of a treatment-invariant representation through domain adversarial losses. \citet{liu2018estimating} implemented propensity score matching in mini-batches as a strategy to mitigate temporal confounding. These methods typically assume treatments with discrete values. However, the efficacy of balancing representations in addressing these issues has recently received attention. Such approaches might only be beneficial in certain conditions, such as with small sample sizes, as suggested by \citet{alaa2018limits}, or involving numerous noisy covariates, as noted by \citet{johansson2022generalization}. \citet{hess2023bayesian} have highlighted that balancing representations might not effectively reduce bias in scenarios with time-varying confounders due to identifiability challenges. Similarly, our supplementary experiments do not support balancing representations as means to address temporal confounding, see \cref{app:sec:temporal_confounding}. Additionally, even in static contexts, these representations face complications due to the invertibility assumption imposed on the learned representations \citep{shalit2017estimating}. For a more comprehensive analysis, we refer the reader to \citet{melnychuk2023bounds} and \citet{curth2021nonparametric}.

\paragraph{Uncertainty quantification in treatment effect estimation.} Various studies have incorporated uncertainty quantification in treatment effect estimations over time. \texttt{TE-CDE} \citep{seedat22b} and \texttt{G-NET} \citep{gnet2021} use Monte Carlo (MC) dropout \citep{gal2016dropout} to approximate the posterior distributions of counterfactual outcomes. However, \citet{hess2023bayesian} note that MC dropout provides poor approximation of a posterior, proposing instead a Bayesian deep learning approach in \texttt{BNCDE} that operates in continuous time. Their findings suggest that deferring treatments with high uncertainty can significantly lower the error in outcome prediction. However, this method only accounts for uncertainty quantification of predictions of single time steps (instead of multiple time steps) and, like previous approaches, is limited to discrete treatments. \texttt{CF-ODE} puts further attention on uncertainty quantification, as it helps in detecting a lack of overlap between treated and non-treated distributions and confounding issues \citep{debrouwer2022predictingimpacttreatmentstime}.

\paragraph{Optimal treatment selection.} Optimal treatment strategy for healthcare, with the goal of maximising clinical outcomes, has been extensively studied \citep{caye2019treatment, dienstmann2015personalizing, zhang2012estimating, murphy2003optimal, robins2004proceedings, moodie2007demystifying}. Different approaches have been studied, including variable selection \citep{lu2013variable, song2015sparse}, multicriteria decision-making methods with conflicting outcomes \citep{bellos2023multicriteria}, optimal treatments selection based on predictive factors \citep{polley2009selecting}, multiple treatments assignment \citep{lou2018optimal}. To the best of our knowledge, previous optimal treatments selection frameworks have not, however, been linked with uncertainty-aware counterfactual estimation in longitudinal data.

\section{Model and Problem Setting}

\paragraph{Problem setting.}
We consider longitudinal datasets, each consisting of $n$ multidimensional patient trajectories. Each trajectory includes: the outcomes ($\bY_t \in \bR^{d_y}$), the treatments ($\bA_t \in \bR^{d_a}$), the observed covariates ($\bX_t \in \bR^{d_x}$). We denote the history of all observed covariates up to time $t$ for patient $i \in \{1,\ldots,n\}$ as $\hH_t^i = \{\bar{\bX}_t^i, \bar{\bA}_{t}^i, \bar{\bY}_t^i\}$, with $\bar{\bX}_t^i := \bX^i_{[0,t]}$. In the following, we omit the $i$ for better readability. This notation follows from \citep{melnychuk22a, Bica2020Estimatingadversariallybalancedrepr}.

We use the potential outcomes frameworks \citep{rubin2005causal}, extended to accommodate time-varying treatments \citep{robins2004proceedings}. Our observational period spans $[0, t]$, and we consider future projections over a time horizon $\tau \geq 0$, within the interval $[t, t+\tau]$. Given a continuous-valued intervention on the treatment ${\mathbf{a}}_{t:t+\tau}=\mathbf{a}_{[t,t+\tau]}$, we are interested in the values of ${\bY}_{t:t+\tau}[{\mathbf{a}}_{t:t+\tau}]$, which represents the potential outcome $\bY$ over the interval $[t, t+\tau]$ following an intervention on ${\mathbf{A}}_{t:t+\tau}$. Generally, works like \citet{Bica2020Estimatingadversariallybalancedrepr} are interested in the point estimate
 $\mathbb{E}[{\bY}_{t:t+\tau}[{\mathbf{a}}_{t:t+\tau}] \given \hh_t]$, corresponding to the future counterfactual outcomes ${\bY}_{t:t+\tau}$ following an intervention ${\mathbf{a}}_{t:t+\tau}$, on a patient with history $\hh_t$. We make the standard assumptions needed for the identification of potential outcomes from observational data: consistency, sequential ignorability, and sequential overlap \citep{lim2018forecasting,robins2004proceedings}.

Similarly to \citet{hess2023bayesian, debrouwer2022predictingimpacttreatmentstime}, our goal is to estimate uncertainty-aware counterfactual trajectories to aid in selecting reliable treatment strategies. Given a desired outcome trajectory ${\by}^*_{t:t+\tau}$, we would like to select the optimal treatment trajectory ${\mathbf{a}}^*_{t:t+\tau}$ such that
\begin{equation*}
\min_{{\mathbf{a}}_{t:t+\tau} \in S({\mathbf{A}}_{t:t+\tau})} \{ \| \mathbb{E} [{\bY}_{t:t+\tau}[{\mathbf{a}}_{t:t+\tau}] \given \hh_t ] - {\by}^*_{t:t+\tau} \|^2 + \lambda \text{Var} [{\bY}_{t:t+\tau}[{\mathbf{a}}_{t:t+\tau}] \given \hh_t ] ]\}\:,
\end{equation*}
where $\lambda$ is the uncertainty weight. Here, $ S(\mathbf{A}_{t:t+\tau}) \subseteq \mathcal{A}$ with $\mathbf{A}_{t:t+\tau} \in \mathcal{A}$ represents the outcome trajectories satisfying a defined constraint.

\paragraph{Uncertainty quantification and treatment selection.} 
We select as counterfactual trajectories estimation methods the baselines: \texttt{CF-ODE}, \texttt{BNCDE}, \texttt{G-NET}, \texttt{CRN}, \texttt{CT}. We will refer to these counterfactual estimation neural network-based models ${G}_{\phi_t}: \bR^{d_a+d_x+d_v} \rightarrow  \bR^{d_y}$, estimating $\mathbb{E}[{\bY}_{t'}[{\mathbf{a}}_{t:t+\tau}] \given \hh_t]$ from ${\mathbf{a}}_{t:t+\tau}$ and $\hh_t$ with $t'\in [t,t+\tau]$. More details on the uncertainty estimation and implementation of models can be found in \cref{app:sec:count_estimation}.

To assess the uncertainty of counterfactual predictions, we employ three model-agnostic techniques: Monte-Carlo (MC) dropout, ensembling, and geometric ensembling, as discussed by \citet{garipov2018losssurfacesmodeconnectivity}. For MC-dropout, we perform eight forward passes, calculating the average  $\hat{\mathbf{\mathbf{\mu}}}_{t:t+\tau}({\mathbf{a}}_t,\hh_t)$ and variance  $\hat{\mathbf{\sigma}}_{t:t+\tau}({\mathbf{a}}_t,\hh_t)$ of the predicted counterfactual outcomes $\hat{\mathbf{y}}_{t:t+\tau} = {G}_{\phi_t}({\mathbf{a}}_{t:t+\tau},\hh_{t:t+\tau})$. Similarly, in the ensembling approach, we utilize a group of eight models to compute these statistics. Geometric ensembling, chosen for its training and computation efficiency, also involves sampling eight neural networks to determine the average and variance of predictions. MC-dropout, due to its superior speed, is our primary method for most treatment selection experiments. Additionally, for models incorporating neural stochastic differential equations (\texttt{BNCDE} and \texttt{CF-ODE}), we conduct eight forward passes to compute both the average and variance of the counterfactual outcome predictions, ensuring consistency in our experimental approach.
We perform treatment selection with the following objective
\begin{equation*}
    \min_{\bar{\mathbf{a}}_{t:t+\tau} \in S({\mathbf{A}}_{t:t+\tau})} \{ \frac{1}{\tau} \sum_{i=1}^{\tau} (\hat{\mathbf{\mathbf{\mu}}}_{t+i}({\mathbf{a}}_{t+i},\hh_{t+i}) - \mathbf{y}_{t+i})^2 + \lambda  \frac{1}{\tau} \sum_{i=1}^{\tau} \hat{\mathbf{\sigma}}_{t+i}({\mathbf{a}}_{t+i},\hh_{t+i})^2 \}\:.
\end{equation*}
More information on the choice of the treatment constraint $S({\mathbf{A}}_{t:t+\tau})$ can be found in \cref{app:sec:constraints_trajectory}.We utilize gradients available through auto-differentiation to perform gradient descent on the loss function using the AdamW optimizer \citep{loshchilov2019decoupled}. 

\paragraph{Evaluation.}
To evaluate how reliably the selected treatments achieve the desired outcomes, we leverage the ground truth dynamics from which the data is simulated. We simulate the counterfactual trajectory ${\mathbf{Y}}_{t:t+\tau}({\mathbf{a}}^*_{t:t+\tau})$ for the selected treatment $\mathbf{a}^*_{t:t+\tau}$ and compare it with the counterfactual estimation ${G}_{\phi_t}({\mathbf{a}}^*_{t:t+\tau},\hh_t)$. We evaluate the reliability of the treatment selection as the root mean squared error between the counterfactual outcome and the desired outcome ${\mathbf{Y}}_{t:t+\tau}({\mathbf{a}}^*_{t:t+\tau})$
\begin{equation*}
\text{RMSE}_{\text{selection}}= \sqrt{\frac{1}{\tau} \sum_{i=1}^{\tau}  (\hat{\mathbf{\mathbf{\mu}}}_{t+i}({\mathbf{a}}^*_{t+i},\hh_{t+i}) - \bar{\mathbf{Y}}_{t+i}({\mathbf{a}}^*_{t+i}))^2}\:.
\end{equation*}

\section{Experiments}

We evaluate our model using different counterfactual estimation baselines, uncertainty quantification methods and treatment constraints on two simulated datasets. More details on the parameters used can be found in \cref{app:sec-experiment-det}.

\subsection{Data}

We utilize two simulated medical datasets focused on the COVID-19 and cardiovascular system to evaluate our model. The use of simulated data is essential as it provides access to true counterfactual trajectories, which are crucial for assessing our method's performance. For all datasets, we simulate 40 time steps of observations in the interval $[0, 30]$ ($t=30$) and consider a prediction window of $[30,40]$, with $\tau = 10$. We generate $\hH_{t+\tau}$  for each patient from randomly sampled initial conditions. We use 1024, 128, and 128 patients as training, validation and test set, respectively. 

\paragraph{COVID-19 dataset.}
We use the synthetic dataset by \citep{qian2021integrating}, which is an extension of the one proposed in \citet{dai2021prototype} following
\begin{align*}
\dot{Z}_1(t) &= k_{dp} Z_1(t) - k_{di} Z_1(t) Z_3(t) - k_{dr} Z_1(t) Z_2(t)\:, \\[-1mm]
\dot{Z}_2(t) &= k_{id} Z_1(t) - k_{io} Z_2(t) + k_{if} Z_1(t) Z_2(t) + \frac{k_{ep} Z_2^{h_p}}{k_{c_p}^{h_p} + Z_2^{h_p}} - k_{d} Z_3(t) Z_2(t)\:, \\[-2mm]
\dot{Z}_3(t) &= k_{im} Z_2(t)\:, \\
\dot{Z}_4(t) &= k_{kel} C(t) - k_{kel} Z_3(t)\:,
\end{align*}
where $\dot{Z}_i(t) = \tfrac{dZ_i(t)}{dt}$ and $ Z_1(t)$, $ Z_2(t)$, $ Z_3(t)$ , $Z_4(t)$ model the disease progression, immune reaction, immunity, and dose, respectively.

To model a clinical setting, we consider multiple treatment cycles, at the beginning of which a clinician can adjust the dose $\theta$ based on intermediate outcomes (5 cycles). Specifically, we consider that the treatment $Z_4(t)$ is increased by 10\% every time the outcome $Z_1(t)$ increases from the previous time step, otherwise it is decreased by 10\%. 

\paragraph{Cardiovascular dataset.}
We use the cardiovascular system (CVS) model proposed in \citet{zenker2007inverse}, modelling the heart, the venous and arterial subsystems and the nervous system control of blood pressure in response to fluid intake. Specifically, we use the same underlying model as \citet{debrouwer2022predictingimpacttreatmentstime,linial2021generative}---a simplification of the one in \citet{zenker2007inverse}
\begin{align*}
    \dot{SV}(t) &=I_{\mathrm{external}}\:,\\
    \dot{P}_a(t) &=C_a^{-1} \Bigl( \frac{P_a(t)-P_v(t)}{R_{\mathrm{TPR}}(S)} -SV \cdot f_{\mathrm{HR}}(S)\Bigr)\:, \\
    \dot{P}_v(t) &=C_v^{-1} ( -C_a\dot{P}_a(t) + I_{\mathrm{external}})\:, \\
    \dot{S}(t) &=\tau_{\mathrm{Baro}}^{-1} \Bigl(1- \frac{1}{1+\exp(-k_{\mathrm{width}}(P_a(t)-P_{a_{\mathrm{set}}})} -S\Bigr)\:,
\end{align*}
 with
\begin{align*}
     R_{\mathrm{TPR}} &= S(t) (R_{{\mathrm{TPR}}_{\max}} - R_{{\mathrm{TPR}}_{\min}}  )+ R_{{\mathrm{TPR}}_{\min}} + R_{{\mathrm{TPR}}_{\mathrm{mod}}}\:, \\
     f_{\mathrm{HR}}(S) &= S(t) ( f_{{\mathrm{HR}}_{\max}} -  f_{{\mathrm{HR}}_{\min}}) +  f_{{\mathrm{HR}}_{\min}}\:, \\
     I_{\mathrm{external}} &= \theta \cdot e^{-\frac{5 - t}{5}}\:.
\end{align*}
Here, $SV$, $P_a$, $P_v$, $S$ represent the cardiac stroke volume, arterial blood pressure, venous blood pressure, and autonomic baroreflex tone, respectively, and $\theta$ represents the dosage.  We consider the setting of a fluid challenge, in which fluid is administered to treat hypotension. In this setting, the outcome is the venous blood pressure $P_v$ and the treatment is a continuous-valued treatment process $I_{\text{external}}$( \citet{debrouwer2022predictingimpacttreatmentstime}). We consider cycles of treatment, sampling a new continuous valued dose at the beginning of each cycle (5 cycles). More details on the parameters, and sampling the initial conditions as well as $\theta$ can be found in \cref{app:sec:dataset-sim}.

\subsection{Results}

\begin{figure}
  \centering
  \includegraphics[width=0.49\textwidth]{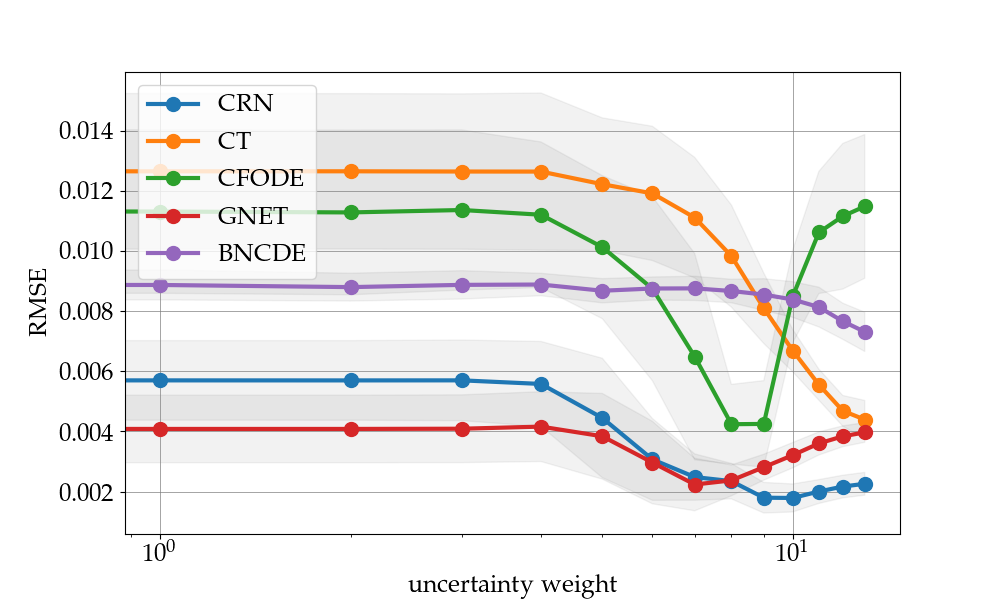}
  \includegraphics[width=0.49\textwidth]{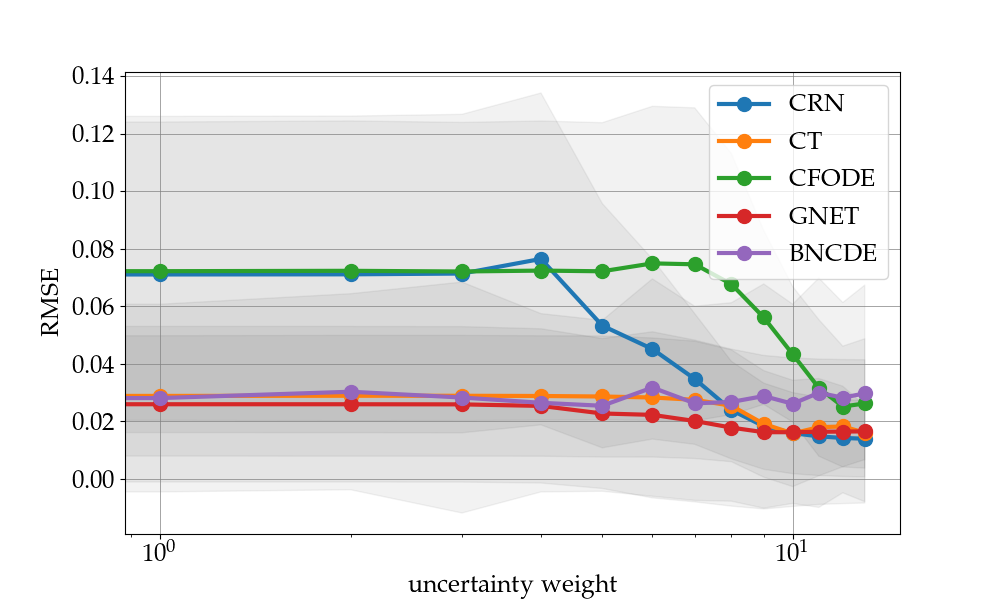}
  \caption{Performance of the baselines \texttt{CRN}, \texttt{CT}, \texttt{CF-ODE}, \texttt{G-NET}, \texttt{BNCDE} when selecting treatments for patients from the cardiovascular (left) and COVID-19 (right) datasets. For most baselines, the $\text{RMSE}_{\text{selection}}$ for the potential outcome compared to the desired outcome decreases, the higher the uncertainty weight in the optimization.}
  \label{fig:metrics-performance}
\end{figure}

\begin{figure}
  \centering
  \includegraphics[width=0.49\textwidth]{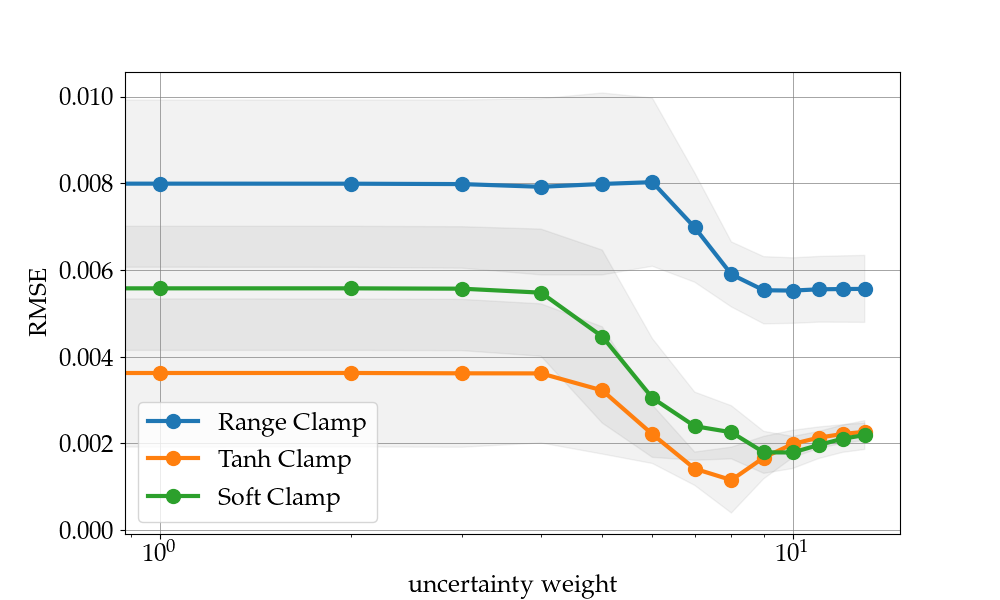}
  \includegraphics[width=0.49\textwidth]{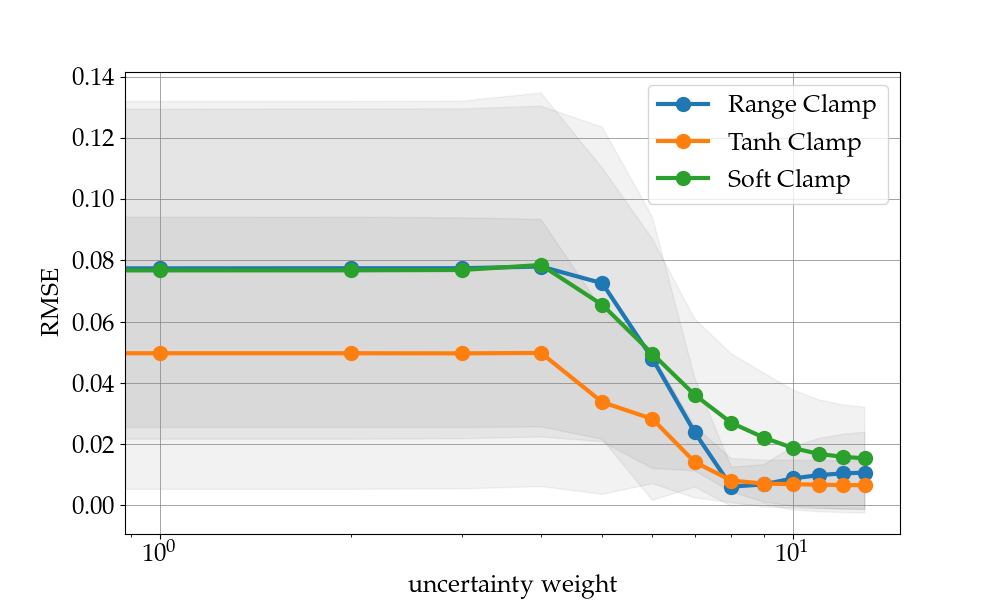}
  \caption{Comparison of clamping constraints for treatment selection, using the \texttt{CRN} as a counterfactual estimator under a range of uncertainty weights. On the left we show results on the cardiovascular dataset and on the right on the COVID-19 dataset. The performance of treatment selection is robust to different treatment constraints choices, with an overall improvement of the performance in treatment selection for higher uncertainty weights.}
  \label{fig:clamping_experiments}
\end{figure}

\begin{figure}
  \centering
  \texttt{CRN} \textbf{on the cardiovascular dataset} \\
  \includegraphics[width=0.49\textwidth]{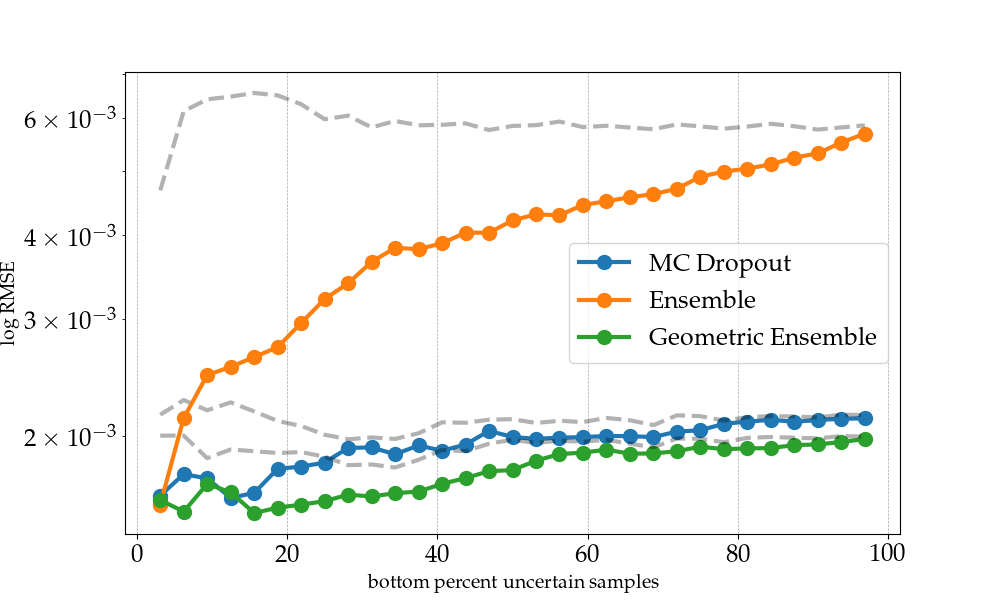}
  \includegraphics[width=0.49\textwidth]{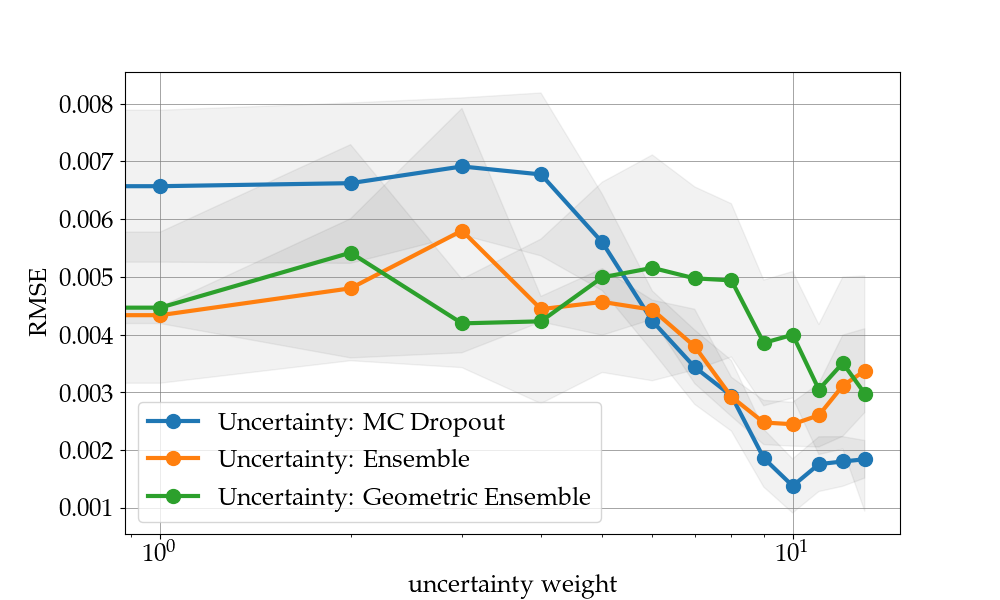}\\[3mm]
  \texttt{CRN} \textbf{on the COVID-19 dataset} \\
  \includegraphics[width=0.49\textwidth]{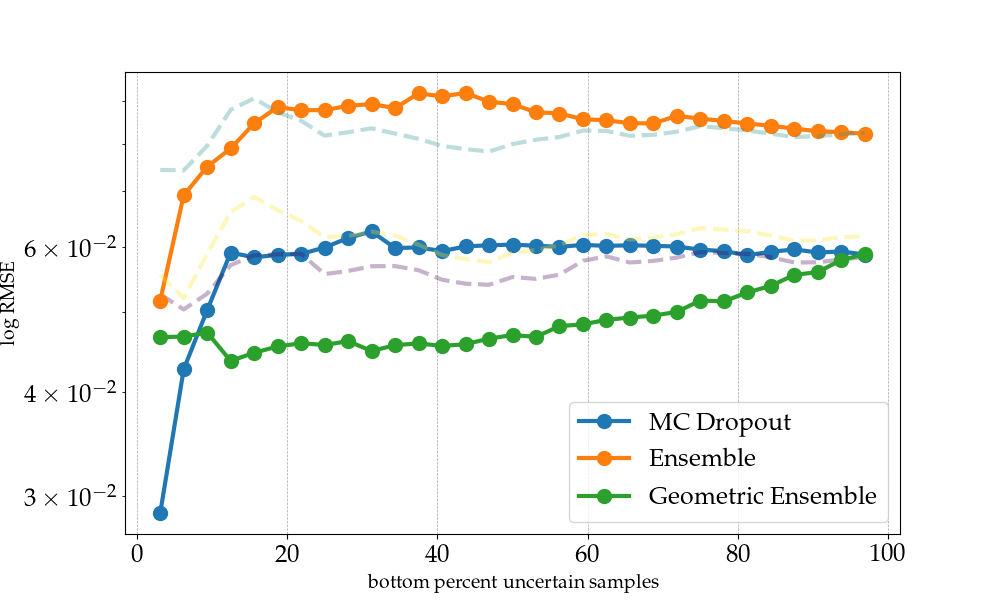}
  \includegraphics[width=0.49\textwidth]{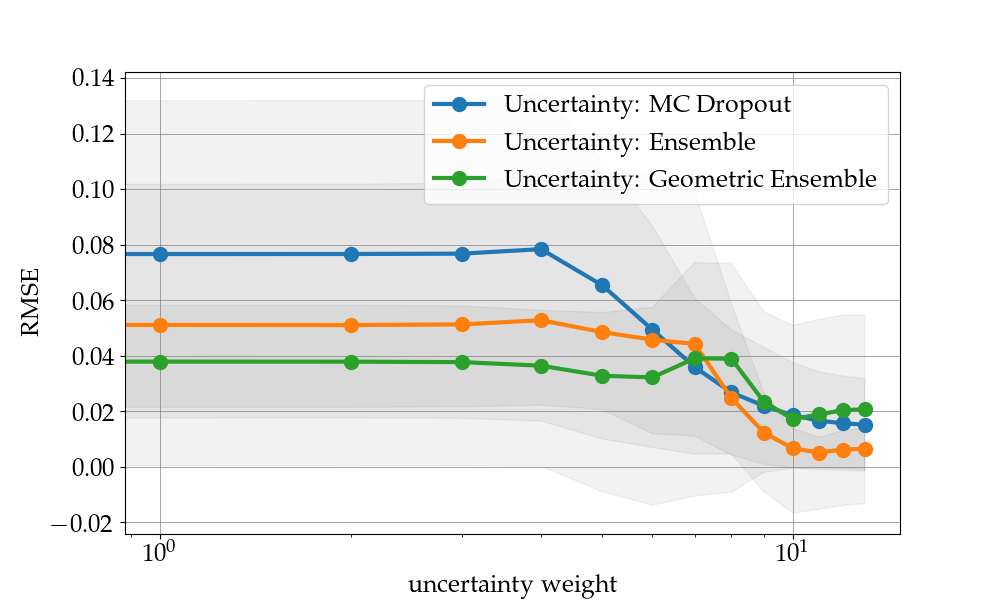}
  \caption{Comparison of the model-agnostic uncertainty quantification methods: MC dropout, ensemble and geometric ensemble methods applied to the \texttt{CRN} on the cardiovascular dataset (upper) and COVID-19 dataset (lower). Performance of treatment selection when varying the uncertainty estimate of the selected samples (left) and varying the uncertainty weight on the whole dataset (right). Selecting the least uncertain samples yields more reliable predictions of the outcome. On the left, we evaluate the models on an increasing percentage of the least uncertain samples (solid line) and compare this to a random subset of the validation data (dashed line). In all uncertainty-quantification methods, treatment selection yields more reliable treatments by increasing the weight on the uncertainty objective.}
  \label{fig:cardiovascular-CRN-uncertainty-quant}
\end{figure}

We evaluate the performance of our method using the cardiovascular and COVID-19 datasets, using different counterfactual estimators. \Cref{fig:metrics-performance} shows how incorporating uncertainty quantification into the optimization objective overall increases the reliability of treatment selection for both datasets. However, in the cardiovascular dataset, certain baselines exhibit a local minimum within the tested range of uncertainty penalties. This phenomenon can be attributed to the dominance of the uncertainty penalty in the treatment selection objective. As the weights of uncertainty in the regularization term increase, the selected treatments deviate from achieving the desired outcomes, resulting in less accurate predictions and an increase in the root mean square error (RMSE). Differently, in the COVID-19 dataset, all the methods show increased performance among all selected uncertainty weights. In \cref{fig:clamping_experiments,fig:cardiovascular-CRN-uncertainty-quant}, the performance of the method is shown to be robust among different treatment constraints and uncertainty metrics respectively using the \texttt{CRN} counterfactual estimation. More details on the clamping approaches can be found in \cref{app:sec:constraints_trajectory}.

\section{Conclusion}

In this work, we propose a method to predict clinical outcomes and optimize treatment effects over time, leveraging uncertainty quantification techniques. Our approach improves on current counterfactual estimation techniques by handling continuous-valued treatment variables -- dosages -- which are common in real-world medical applications. The results from testing on two synthetic datasets show that the proposed method could be applied alongside counterfactual estimation techniques to improve the treatment selection process.

Our method's ability to prioritize treatments that yield highly certain outcomes within a controlled treatment cost framework could be beneficial for personalized medicine: it ensures that the selected treatments are not only cost-effective but also minimize the potential risks associated with high-dose treatments. The added uncertainty component in the treatment selection process enhances decision-making, leading to more reliable patient care outcomes. The strength of our method lies in its robust performance across a range of uncertainty quantification models, counterfactual estimation techniques, and treatment parameters. Put together, its cost-effectiveness, reliability and robustness makes our proposed method well-suited to a variety of medical contexts, offering practitioners the flexibility to identify the optimal treatment approach. Future works could extent the method with additional uncertainty quantification methods and treatment constraints.

\newpage
\begin{acksection}
We thank Valentyn Melnychuk and Konstantin Heß for helpful discussions.
The authors gratefully acknowledge the computational and data resources provided by the Leibniz Supercomputing Centre (\url{www.lrz.de}).
\end{acksection}

\bibliography{neurips_2024}
\bibliographystyle{plainnat}

\newpage
\appendix

\section*{Appendix}

\section{Counterfactual Estimation Baselines}
\label{app:sec:count_estimation}

We consider different counterfactual estimation methods in our models and compare their performance. In our implementations, we modify all methods to handle continuous-valued treatments.

\begin{itemize}[leftmargin=*]
    \item \textbf{BNCDE} \citep{hess2023bayesian}: The BNCDE method combines neural stochastic differential equations for Bayesian uncertainty quantification with neural controlled differential equations for continuous-time treatment effect estimation. Neural stochastic differential equations enable tractable variational Bayesian inference, allowing for the parameterization of posterior distributions of network weights. We build on the implementation by \citet{hess2023bayesian},  extending the method to continuous treatments. At inference time, we allow multi-step prediction by predicting outcomes from latents at several time steps, instead of the final time step. 
    \item \textbf{G-NET} \citep{gnet2021}: G-Net is a sequential deep learning framework for counterfactual prediction based on g-computation. It leverages recurrent neural networks (RNNs) to model time-varying covariates and treatments in complex longitudinal data settings. G-Net estimates the effects of dynamic treatment strategies by simulating patient trajectories under alternative treatments. By employing RNNs, G-Net captures both temporal dependencies and non-linearities in the data, providing more accurate predictions compared to classical models, particularly for counterfactual predictions in clinical and simulated environments like cardiovascular simulations and tumor growth datasets. Our implementation of the G-NET uses GRU-RNNs and a common head for predicting outcomes and covariates.
    \item \textbf{CT}: The Causal Transformer (CT) is designed to estimate counterfactual outcomes over time by capturing long-range dependencies in time-series data, such as patient trajectories. The model incorporates three parallel transformer subnetworks that process different input sequences: time-varying covariates, past treatments, and past outcomes. These are fused using cross-attention to create a balanced representation. Additionally, CT uses Counterfactual Domain Confusion (CDC) loss to minimize confounding bias and produce reliable counterfactual predictions. We replace the CDC loss with the HSIC as a balancing criterion that is suitable for continuous-valued treatments (see Section \ldots for more details). 
    \item \textbf{CRN}: The Counterfactual Recurrent Network (CRN) is a sequence-to-sequence model designed for estimating treatment effects over time. It employs an encoder-decoder architecture where the encoder processes patient history, including time-varying covariates and past treatments, to generate a treatment-invariant representation. This representation is refined through adversarial training to minimize the influence of confounding variables. The decoder uses this balanced representation to predict outcomes under alternative treatment plans, allowing for reliable counterfactual estimations and treatment recommendations over time.  We use HSIC for balancing representation loss for the considered continuous treatment setting (see Section \ldots for more details) and GRU RNNs. 
    \item \textbf{CF-ODE}: The Counterfactual ODE (CF-ODE) method models treatment effects over time and their epistemic uncertainty using an NSDE.  The model encodes the observed time series into a latent space via a GRU RNN, and then integrates the hidden state forward in time using a treatment-specific processes within a NSDE. We extend this model to continuous-valued treatments by replacing the binary treatment-specific processes with an encoding of continuous-valued treatments. Specifically, we add the final hidden state of a GRU RNN processing the treatments to the hidden state to be integrated.  
\end{itemize}
In our experiments, we apply Reversible Instance Normalization (RevIN) \citep{kim2021reversible} to all settings, leading the input to be normalized at the start and denormalized at the output, ensuring consistency in data distributions and reducing the impact of distribution shifts between training and test sets. 

\section{Temporal Confounding and Balancing Representation} 
\label{app:sec:temporal_confounding}

When considering counterfactual estimation, time-varying confounding might lead to a bias in the treatment assignment $A_{0:t}$ in the observational distribution. This is usually referred to as confounding bias. 
\citet{Bica2020Estimatingadversariallybalancedrepr, melnychuk22a} use an adversarial objective to produce a sequence of balanced representations, which are simultaneously predictive of the outcome but non-predictive of the current treatment assignment, resulting in a treatment-invariant representation of the patient history $\Phi(\hh_t^i)$, breaking the association between the treatment history and the treatment assignment. \citet{Bica2020Estimatingadversariallybalancedrepr, melnychuk22a} show that these representations can remove the bias from time-varying confounders. More specifically, they learn a representation such that $P(\Phi(\hh_t) \given \bA_t = \mathbf{a}^{(k)}) \quad \text{is equal for all} \quad k = 1, 2, \ldots, K$ treatments.
The current adversarial-learning based methods of \citet{Bica2020Estimatingadversariallybalancedrepr, melnychuk22a} are built for discrete treatment variables. We extend this balancing objective to to learn a treatment invariant representation $\Phi(\hH_t)$ to continuous-valued treatments. Instead of an adversarial loss with an auxiliary treatment classifier head, we encourage treatment invariant representations by minimizing the Hilbert-Schmidt independence criterion between past treatments and a representation of the trajectory $\text{HSIC}(\bA_{t}, \Phi(\hH_t))$. 
In contrast to adversarial losses, the HSIC is non-parametric and robust to class imbalance in treatments. 

\begin{table}
\centering
\caption{Parameters for Cardiovascular Simulation}
\label{tab:params-cardio-updated}
\begin{tabular}{lll}
\toprule
\textbf{Parameter} & \textbf{Description} & \textbf{Value} \\
\midrule
$f_{hr_{\text{max}}}$ & Maximum heart rate scaling factor & 3.0 \\
$f_{hr_{\text{min}}}$ & Minimum heart rate scaling factor & 0.6666 \\
$r_{tpr_{\text{max}}}$ & Maximum total peripheral resistance & 2.134 \\
$r_{tpr_{\text{min}}}$ & Minimum total peripheral resistance & 0.5335 \\
$r_{tpr_{\text{mod}}}$ & Modifier for total peripheral resistance & 0.0 \\
$sv_{\text{mod}}$ & Stroke volume modification due to intervention (matches $\theta$) & 0.001 \\
$ca$ & Arterial compliance & 4.0 \\
$cv$ & Venous compliance & 111.0 \\
$k_{\text{width}}$ & Width parameter for sigmoid function & 0.1838 \\
$p_{\text{aset}}$ & Setpoint arterial pressure & 70 \\
$\tau$ & Time constant for autonomic response & 20 \\
$p_{0\text{lv}}$ & Left ventricular preload pressure & 2.03 \\
$r_{\text{valve}}$ & Valve resistance & 0.0025 \\
$k_{\text{elv}}$ & Left ventricular elastance coefficient & 0.066 \\
$v_{ed0}$ & Initial end-diastolic volume & 7.14 \\
$T_{\text{sys}}$ & Systole duration & 0.2666 \\
$cprsw_{\text{max}}$ & Maximum pressure-volume relation (cprsw) & 103.8 \\
$cprsw_{\text{min}}$ & Minimum pressure-volume relation (cprsw) & 25.9 \\
\bottomrule
\end{tabular}
\end{table}

\begin{table}
\centering
\caption{Parameters for COVID-19 Dataset Simulation}
\label{tab:params-covid-updated}
\begin{tabular}{lll}
\toprule
\textbf{Parameter} & \textbf{Description} & \textbf{Value} \\
\midrule
$hill_{\text{cure}}$ & Hill coefficient for cure response & 2.0 \\
$h_p$ & Hill coefficient for pathogen response & 2.0 \\
$k_{c_p}$ & Half-maximal effective concentration for pathogen & 1.0 \\
$k_{ep}$ & Maximum effect of pathogen response & 1.0 \\
$k_{d}$ & Drug effect rate (pathogen-related) & 1.0 \\
$k_{dp}$ & Disease progression rate & 1.0 \\
$k_{dr}$ & Immunity-driven disease removal rate & 1.0 \\
$k_{di}$ & Immune reaction to disease & 1.0 \\
$k_{id}$ & Disease-induced immune response activation rate & 1.0 \\
$k_{if}$ & Immune feedback rate & 1.0 \\
$k_{io}$ & Immune response decay rate & 1.0 \\
$k_{im}$ & Immunity buildup rate & 1.0 \\
$k_{kel}$ & Drug elimination rate & 1.0 \\
\bottomrule
\end{tabular}
\end{table}

We assess the performance of balancing representations on continuous-valued treatments. To model different levels of confounding, at the beginning of each treatment cycle $i$ we sample doses $\theta$ from a beta distribution parameterized as following \citet{bica2020estimatinggans}
\begin{equation*}
\theta_{i+1} \sim \text{Beta} (\alpha, \beta), \quad \text{where } \beta = \frac{\alpha - 1}{d_w^{i+1}} + 2 - \alpha \:.
\end{equation*}
For both data simulations, we set $d_w^{0} = \mathbf{A}_{0}$.

For the COVID-19 data simulation, we specify $d_w^{i+1}$ as
\begin{equation}
    d_w^{i+1} = 
    \begin{cases} 
    d_w^{i} \times 1.1 & \text{if } y_0^{i+1} \geq y_0^{i}\:, \\
    d_w^{i} \times 0.9 & \text{if } y_t^{i+1} \leq y_0^{i}\:.
    \end{cases}
\end{equation}
For the cardiovascular data simulation, we do not specify $d_w^{i+1}$, but instead set $d_w^{i+1} = d_w^{i} = \ldots = d_w^{1} = \mathbf{A_0}$.
When $\alpha = 1$, the distribution becomes uniform, and so $\theta$ is independent of previous outcomes, treatments and covariates---preventing confounding entirely. Increasing $\alpha$ yields stronger confounding.

\begin{figure}
  \centering
  \includegraphics[width=0.8\textwidth]{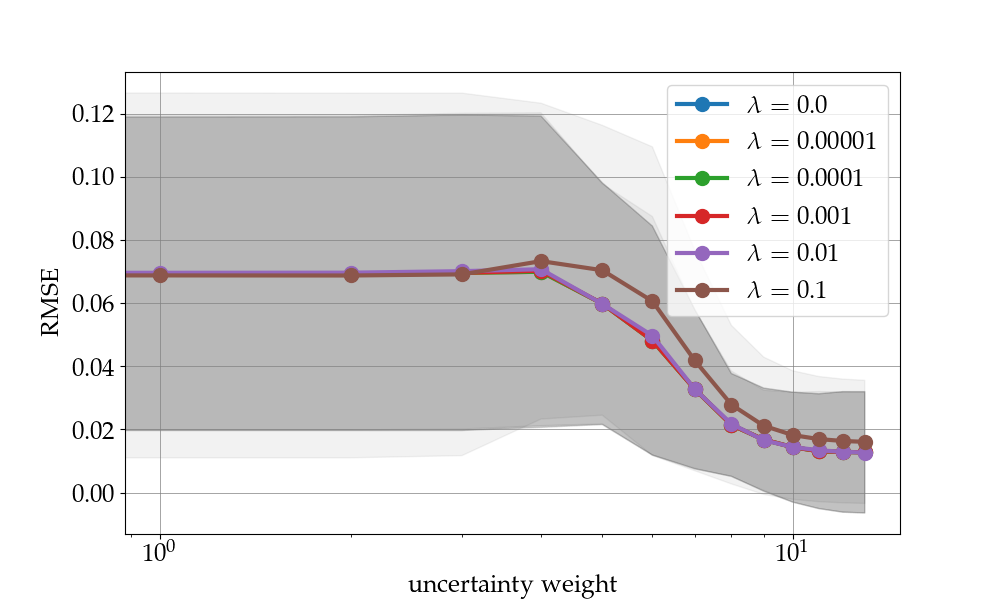}
  \caption{Encouraging a balancing representations with increasing weight on the HSIC objective has a negligible effect on reliable counterfactual estimation compared to the uncertainty objective}
  \label{fig:confounding}
\end{figure}

To check if balancing representations help in our settings, we set $\alpha=2$ and up-weigh the balancing objective in our loss, in factors of $\lambda$. In line with the literature \citet{alaa2018limits}, we find no benefit for $\lambda \geq 0$ in \cref{fig:confounding}.

\section{Dataset Simulation}
\label{app:sec:dataset-sim}

We sample the initial conditions of the cardiovascular dataset from the following distributions:
\begin{align*}
   \text{SV} &\sim \text{Unif}(0.9, 1.0)\:, \\
   P_a &\sim \text{Unif}(0.75, 0.85)\:,\\
   P_v &\sim \text{Unif}(0.3, 0.7)\:,\\
   s &\sim \text{Unif}(0.15, 0.25)\:.
\end{align*}
and for the COVID-19 dataset:
\begin{equation*}
    Z_i \sim \text{Exp}(0.01)\quad \text{for } i \in \{1, 2, 3, 4\}\:.
\end{equation*}

A more detailed overview of the parameters for the cardiovascular and COVID-19 datasets simulation can be found in \cref{tab:params-cardio-updated} and \cref{tab:params-covid-updated} respectively.

\section{Experiments}

\subsection{Constraints on Treatment Trajectories}
\label{app:sec:constraints_trajectory}

To ensure optimal treatment administration within a clinically acceptable framework, we use different constraints on the dosage trajectory ${\bA}_{t:t+\tau}$  such that $ S({\mathbf{A}}_{t:t+\tau}) \subseteq \mathcal{A}$. This is achieved by applying a mapping $v: \mathbb{R}^{\tau \times d_a} \to S(\mathbf{A}_{t:t+\tau})$, where $v(\mathbf{A}_{t:t+\tau}) \in S(\mathbf{\bA}_{t:t+\tau})$.

\begin{itemize}
    \item \textbf{Range Clamping}: Each dosage is adjusted to remain within a specified range, mitigating the risk of excessively high or low doses. This is implemented by centering and clamping the dosage values within the interval $[a, b]$
    \begin{equation*}
        v_{range}(\mathbf{A}_{t:t+\tau}) = \min (\max (\mathbf{A}_{t:t+\tau} - \text{Mean}(\mathbf{A}_{t:t+\tau}), a  ), b  )\:.
    \end{equation*}

    \item \textbf{Soft Clamping}: To discourage extreme dosages while improving optimization, a soft clamping approach is applied, which pulls values towards a central range defined by $\beta$
    \begin{equation*}
        v_{soft}(A_{t:t+\tau}) = 
        \begin{cases} 
            \alpha \mathbf{A}_{t:t+\tau} & \text{if } \mathbf{A}_{t:t+\tau} > \beta\:, \\
            \mathbf{A}_{t:t+\tau} & \text{if } -\beta \leq \mathbf{A}_{t:t+\tau} \leq \beta\:, \\
            \alpha \mathbf{A}_{t:t+\tau} & \text{if } \mathbf{A}_{t:t+\tau} < -\beta \:.
        \end{cases}
    \end{equation*}

    \item \textbf{$\tanh$ Clamping}: To limit dosages within a range $(-\beta, \beta)$  while providing a continuous function, we apply the $\tanh$ clamping method
    \begin{equation*}
        v_{tanh}(\mathbf{A}_{t:t+\tau}) = \beta \cdot \tanh(\mathbf{A}_{t:t+\tau})\:.
    \end{equation*}
\end{itemize}
These constraints allow flexibility within defined clinical parameters for determining dosage administration constraints. These seemed the most realistic and interesting settings, however changes to these constraints can be made based on specific clinical needs or treatment goals.

\begin{table}
\centering
\caption{Parameters for Treatment Selection}
\label{tab:params-treatment-selection}
\begin{tabular}{ll}
\toprule
\textbf{Parameter} & \textbf{Value} \\
\midrule
Uncertainty Weights & $0$, $10^{-5}$, $10^{-4}$, $10^{-3}$, $10^{-2}$, $0.0625$, $0.125$, $0.25$, $0.5$, $1$, $2$, $4$, $8$, $16$ \\
MSE Weight & $0.02$ \\
Constraints & Soft clamp, $\beta = 4$, $\alpha = 0.01$ \\
Optimizer & AdamW \citep{loshchilov2019decoupled} \\
Optimization Steps & $50$ \\
Learning Rate & $0.1$ \\
Replicates & $6$ \\
\bottomrule
\end{tabular}
\end{table}

\subsection{Experiment Details}
\label{app:sec-experiment-det}
We present the parameters used for treatment selection in \cref{tab:params-treatment-selection}.

\begin{figure}
  \centering
    \includegraphics[width=0.49\textwidth]{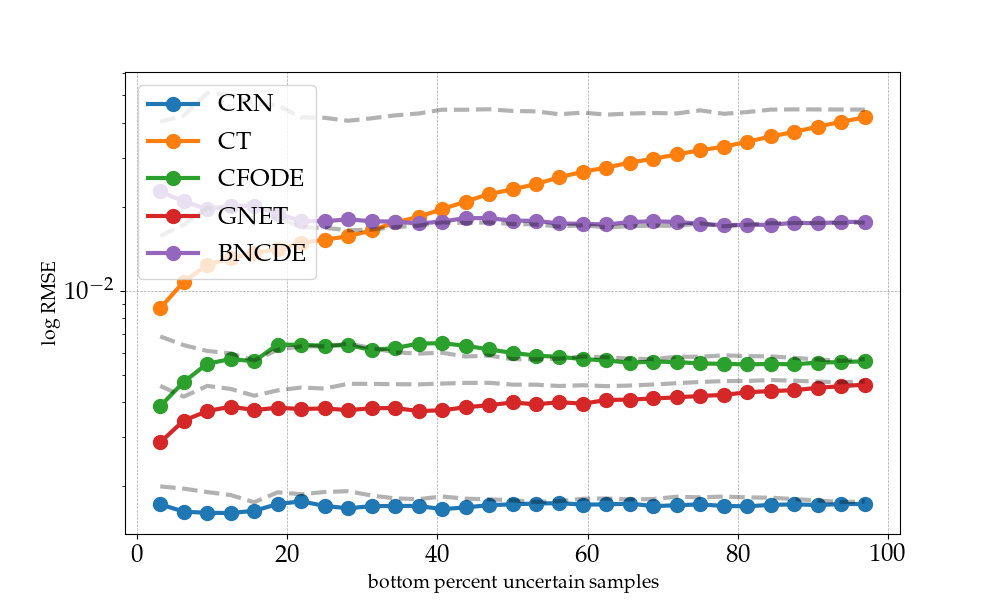}
    \includegraphics[width=0.49\textwidth]{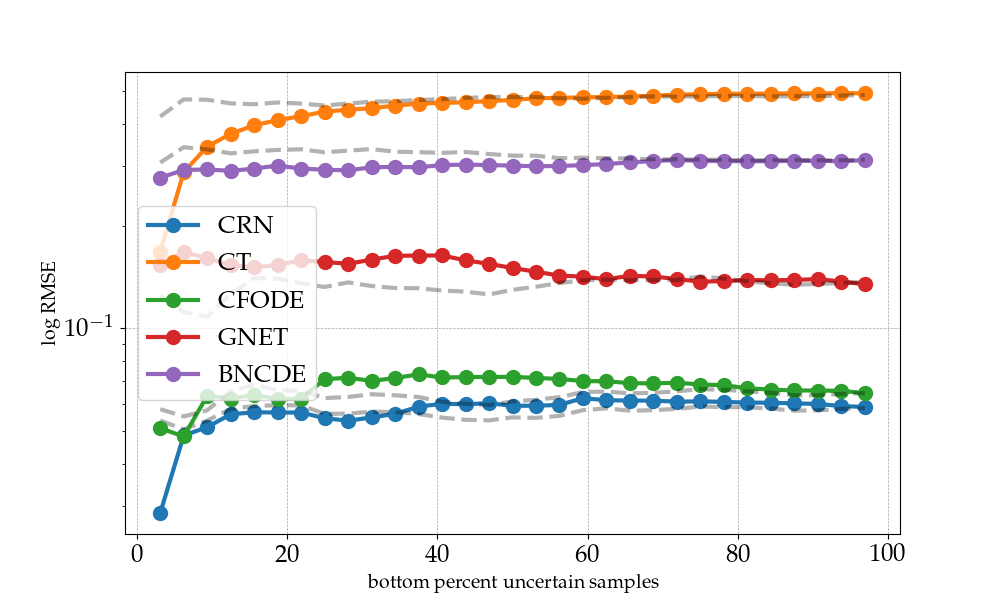}
  \caption{Performance comparison of the baseline models \texttt{CRN}, \texttt{CT}, \texttt{CF-ODE}, \texttt{G-NET}, \texttt{BNCDE} in treatments selection for cardiovascular (left) and COVID-19 (right) datasets. This comparison includes varying percentages of the most uncertain samples alongside a random subset of the validation data. The trends observed across different baselines are relatively stable, showing overall improved performance with increased certainty in the sample selection.}
  \label{fig:uncertainty-sample-performance}
\end{figure}

\begin{figure}
  \centering
  \texttt{G-NET} \textbf{on the cardiovascular dataset} \\
    \includegraphics[width=0.49\textwidth]{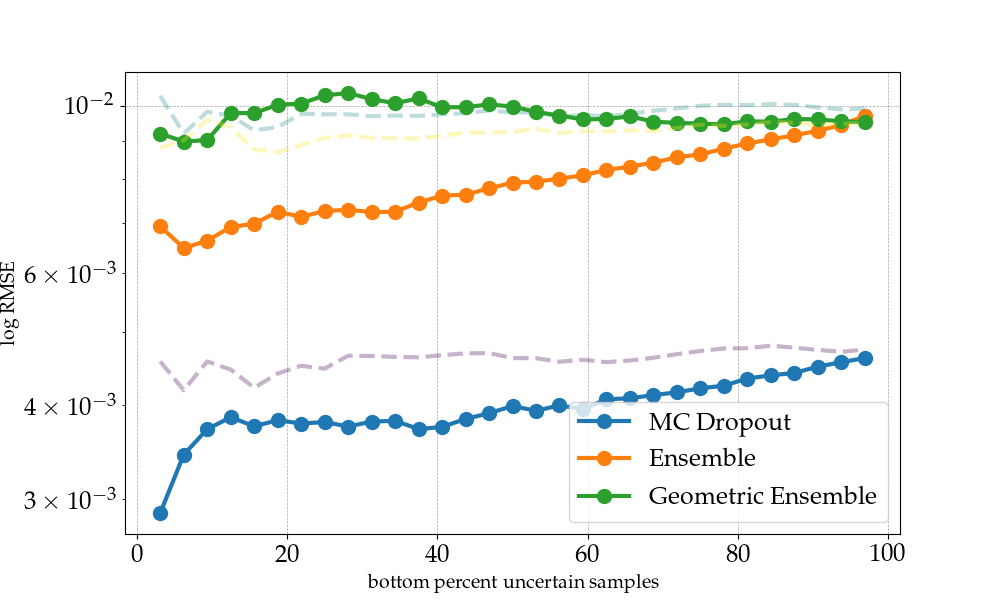}
    \includegraphics[width=0.49\textwidth]{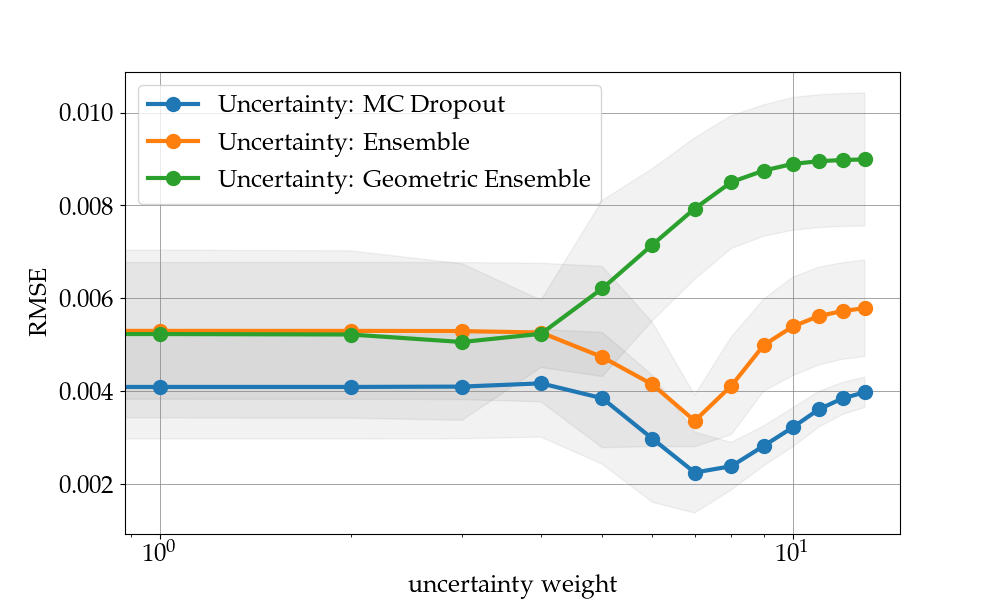}\\[3mm]
  \texttt{G-NET} \textbf{on the cardiovascular dataset} \\
    \includegraphics[width=0.49\textwidth]{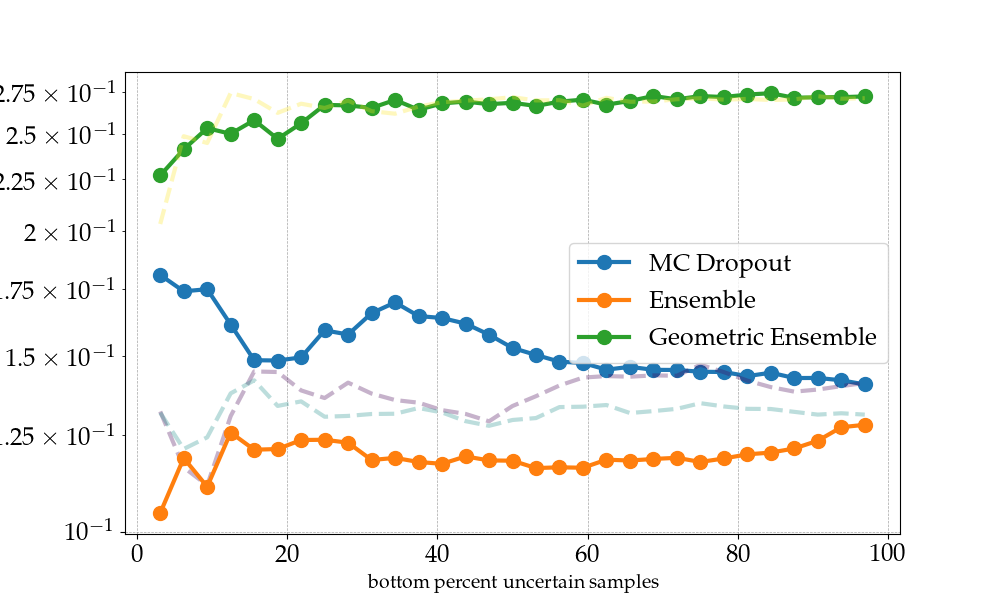}
    \includegraphics[width=0.49\textwidth]{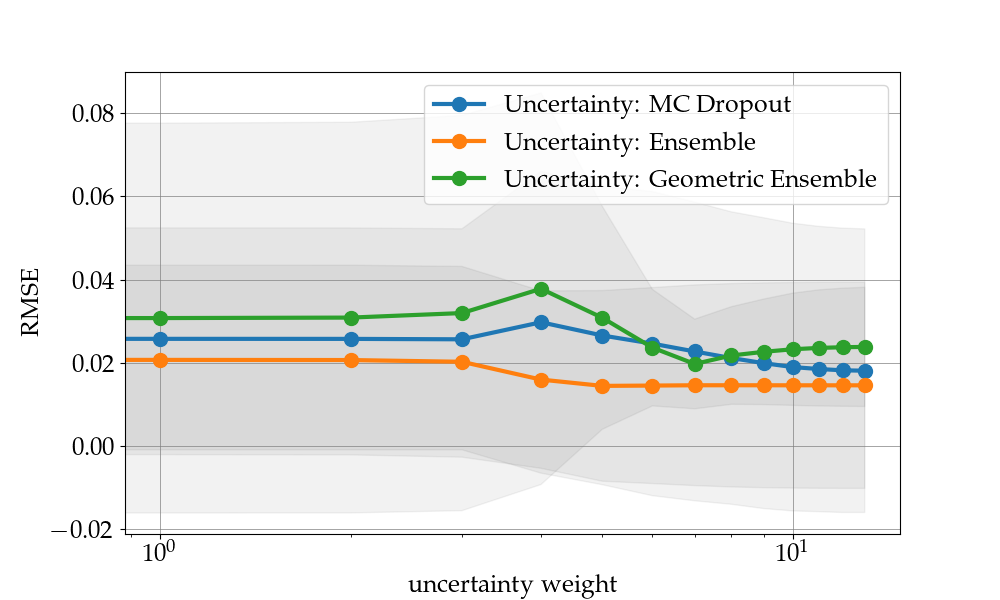}
  \caption{Comparison of the model-agnostic uncertainty quantification MC dropout, ensemble and geometric ensemble methods applied to the \texttt{G-NET} on the cardiovascular dataset (up) and COVID-19 dataset (low). Performance of treatment selection by varying the certainty of the selected samples (left) and by varying uncertainty weight on the whole dataset (right). Selecting the least uncertain samples yields more reliable predictions of the outcome. We evaluate the models on an increasing percentage of the least uncertain samples together with a random subset of the validation data.}
  \label{fig:cardiovascular-gnet-uncertainty-quant}
\end{figure}

\subsection{Additional Experiments}

We also show more results with the different counterfactual estimation baselines in \cref{fig:uncertainty-sample-performance} and we assess the performance of the method with different uncertainty quantification methods with an additional model -- the \texttt{G-NET} -- in \cref{fig:cardiovascular-gnet-uncertainty-quant}.

\end{document}